%% file: main.tex
\documentclass[10pt,twocolumn,letterpaper]{article}

\usepackage[pagenumbers]{cvpr} % To force page numbers, e.g. for an arXiv version

\usepackage{makecell}
\usepackage{graphicx}
\usepackage{subcaption}
\usepackage{graphicx}
\usepackage{multirow}
\usepackage{array}
\usepackage{tabularx}
\usepackage{makecell}
\usepackage{xspace}

\usepackage{pifont}

\definecolor{cvprblue}{rgb}{0.21,0.49,0.74}
\usepackage[pagebackref,breaklinks,colorlinks,allcolors=cvprblue]{hyperref}

\usepackage{tcolorbox}

\newcommand{\cmark}{\ding{51}} % checkmark
\newcommand{\xmark}{\ding{55}} % cross
\tcbset{
  mybox/.style={
    colback=blue!5,
    boxrule=1pt, % Change this value to set the border thickness
    boxsep=2pt,
    left=0pt, % Increased left margin
    right=0pt, % Increased right margin
    top=10pt, % Increased top margin
    bottom=10pt, % Increased bottom margin
    oversize=2pt,
    arc=4pt, % Set this value to adjust the radius of the corners
    before skip=10pt, % Increased space before the box
    after skip=10pt, % Increased space after the box
    baseline=3.5ex % Adjust this value to set line spacing
  }
}

\def\paperID{2337} % *** Enter the Paper ID here
\def\confName{CVPR}
\def\confYear{2025}

\newcommand{\method}{\textsc{ProVision}\xspace}
\newcommand{\data}{\textsc{ProVision-10M}\xspace}

\title{\method: Programmatically Scaling Vision-centric Instruction Data\\ for Multimodal Language Models}

\author{Jieyu Zhang$^1$, Le Xue$^2$, Linxin Song$^3$, Jun Wang$^2$, Weikai Huang$^1$, Manli Shu$^2$, An Yan$^2$, Zixian Ma$^1$, \\Juan Carlos Niebles$^2$, Silvio Savarese$^2$, Caiming Xiong$^2$, Zeyuan Chen$^2$, Ranjay Krishna$^1$\thanks{Corresponding authors.}, Ran Xu$^2$\footnotemark[1]
\vspace{2mm} \\  \vspace{2mm}
$^1$University of Washington, $^2$Salesforce Research, $^3$University of Southern California\\
\includegraphics[height=4mm]{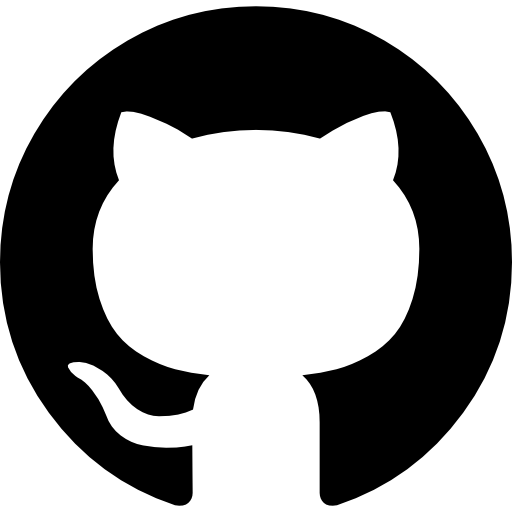} Code: \url{https://github.com/JieyuZ2/ProVision}\\
\raisebox{-1mm}{\includegraphics[height=5mm]{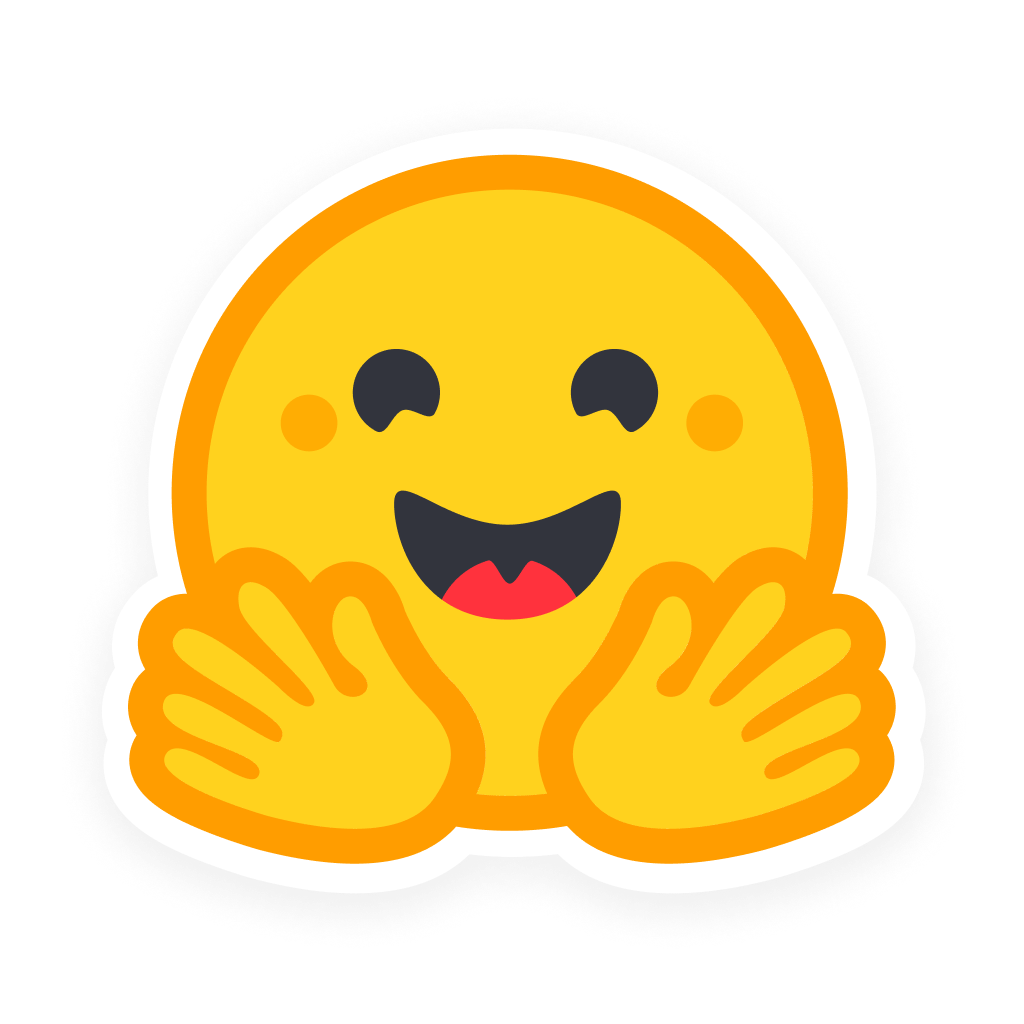}} Dataset: \url{https://huggingface.co/datasets/Salesforce/ProVision-10M}
}

\newcommand{\vgs}{\textbf{VG-S}\xspace}
\newcommand{\vgm}{\textbf{VG-M}\xspace}
\newcommand{\dcs}{\textbf{DC-S}\xspace}
\newcommand{\dcm}{\textbf{DC-M}\xspace}

\begin{document}
\maketitle
\input{sec/0_abstract}    
\input{sec/1_intro}

\input{sec/2_method}

\input{sec/3_experiments}

\input{sec/4_related}

\input{sec/5_conclusion}

{
    \small
    \bibliographystyle{ieeenat_fullname}
    \bibliography{main}
}

\input{sec/X_suppl}

\end{document}

%% file: sec/0_abstract.tex
\begin{abstract}

With the rise of multimodal applications, instruction data has become critical for training multimodal language models capable of understanding complex image-based queries. Existing practices rely on powerful but costly large language models (LLMs) or multimodal language models (MLMs) to produce instruction data. These are often prone to hallucinations, licensing issues and the generation process is often hard to scale and interpret. In this work, we present a programmatic approach that employs scene graphs as symbolic representations of images and human-written programs to systematically synthesize vision-centric instruction data. Our approach ensures the interpretability and controllability of the data generation process and scales efficiently while maintaining factual accuracy. 
By implementing a suite of 24 single-image, 14 multi-image instruction generators, and a scene graph generation pipeline, we build a scalable, cost-effective system: \method which produces diverse question-answer pairs concerning objects, attributes, relations, depth, etc., for any given image. Applied to Visual Genome and DataComp datasets, we generate over 10 million instruction data points, \data, and leverage them in both pertaining and instruction tuning stages of MLMs.
When adopted in the instruction tuning stage, our single-image instruction data yields up to a 7\% improvement on the 2D split and 8\% on the 3D split of CVBench, along with a 3\% increase in performance on QBench2, RealWorldQA, and MMMU.
Our multi-image instruction data leads to an 8\% improvement on Mantis-Eval.
Incorporation of our data in both pre-training and fine-tuning stages of xGen-MM-4B leads to an averaged improvement of 1.6\% across 11 benchmarks.

\end{abstract}

%% file: sec/1_intro.tex
\section{Introduction}
\label{sec:intro}

The success of Multimodal Language Models (MLMs) such as LLaVA and InstructBLIP has been largely built upon the availability of multimodal data~\cite{datacomp,lin2014microsoft,SoricutDSG18}, and visual instruction data~\cite{liu2024visual, dai2024instructblip}. In particular, visual instruction data is key to enable MLMs to follow the instruction and respond to user questions about input images effectively.
To gather visual instruction data, existing practice mainly relies on powerful Large Language Models (LLMs) or MLMs to generate such data samples~\cite{liu2024visual,wang2023instruct4v,liu2024mminstructgeneratedvisualinstructions,liu2024mminstruct}.
While effective, it does come with certain limitations.
First, the generation process remains largely a black-box mechanism, making it difficult to interpret the process and control or customize outputs precisely.
Second, even the most advanced LLMs or MLMs are still prone to hallucination~\cite{cui2023holistic, ji2023survey, snyder2024early, huang2023survey, zhao2023beyond, sadat2023delucionqa, vakharia2023don}, generating content that can be factually inaccurate, which undermines the reliability of the resulting visual data and is typically hard to detect and correct ex post.
Third, the reliance on powerful LLMs or MLMs might hinder the scalability of the data generation process due to the potential cost (such as API usage costs) and entail license constraints that prevent the use of generated data for model training~\cite{openai_terms}.

\begin{figure*}[t]
    \centering
    \includegraphics[width=\linewidth,page=1]{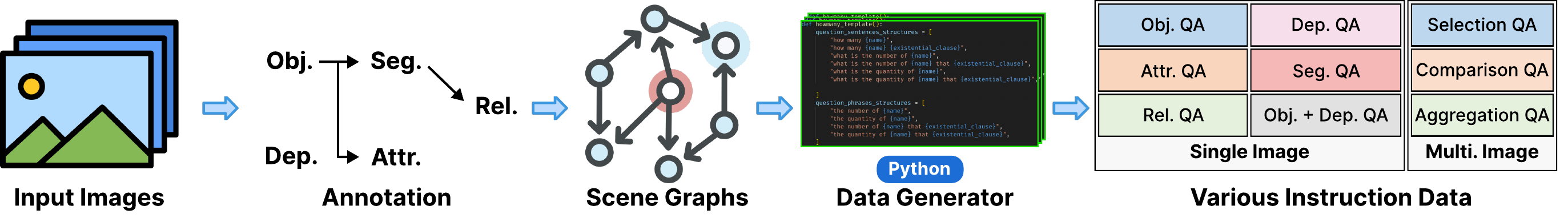}
    \caption{\label{fig:pipeline} \textbf{Overview of \method system}. It can generate a scene graph for any image, enabling programs to synthesize instruction data procedurally from the scene graph. If the scene graph exists, it directly generates instruction data based on it.}
\end{figure*}

In this work, we explore a complementary approach for programmatically generating visual instruction data.
To enable programmatic generation, we leverage \emph{scene graphs}~\cite{vg} as a structured representation of image semantics. We develop programs to systematically generate visual instruction data using automatically extracted scene graph representations from images. 
In a scene graph, each object is represented as a node, where the attributes of the object—such as color, size, or materials—are assigned directly to that node. The relationships or interactions between these objects are depicted as directed edges connecting the corresponding nodes. 
Given a scene graph, a program can generate questions like ``How many red objects are there?'' alongside the ground truth answer. Extending this to multiple scene graphs, another program can create comparative questions such as ``Which image has the most red objects?'', facilitating multi-image instruction data.
Notably, this programmatic approach allows us to readily produce both multiple-choice and short-answer questions.

This approach helps to mitigate the limitations of existing LLM/MLM-driven approaches.
First, assuming correct scene graphs, combined with human-written programs, introduces transparency and interpretability into the data creation process, and enables a more controlled, customizable output generation, eliminating much of the unpredictability that arises in end-to-end models.
Second, programs produce instructions devoid of hallucinations as long as the underlying scene graphs are accurate. 
Rather than relying on probabilistic outputs from LLMs, our instructions are grounded in the explicit information captured within scene graphs.
Furthermore, by shifting the workload from powerful LLMs to programmatic generation, this method enables a scalable and cost-effective solution for data creation. 
Finally, this approach avoids the licensing constraints associated with LLM- or MLM-generated data, as scene graphs and custom programs do not have to involve proprietary model outputs.

We develop \method, a scalable programmatic system as shown in Figure~\ref{fig:pipeline} and introduce \data, a 10 million instructional dataset. \data is created from Visual Genome~\cite{vg} and DataComp~\cite{datacomp}.
We implement a total of 24 single-image instruction data generator programs and 14 multi-image instruction data generator programs. These programs cover a diverse range of image-based queries, addressing aspects such as objects, attributes, relations, segmentation, and depth.
For Visual Genome, we leverage its manually annotated scene graphs.

Not all datasets come with human-annotated scene graphs. For example, DataComp images do not have ground truth scene graphs like Visual Genome does.
To scale \method, we automatically generate scene graphs using a pipeline consisting of state-of-the-art models for object/relation detection~\cite{cheng2024yolo,Robin}, image segmentation~\cite{sam2}, depth estimation~\cite{yang2024depth}, \etc. 
The scene graph pipeline automatically generates a scene graph for any image, which the instruction data generator programs can then utilize to produce both single-image and multi-image instructional data.

Using \data, we train various MLMs, experimenting with both pre-training and instruction tuning stages, varying data scales, and different data formats (multiple choice vs. short answer) to evaluate the effectiveness of our generated instruction data.
Specifically, our single-image instruction data leads to at most 7\% and 8\% improvement of CVBench's 2D and 3D splits, around 3\% improvement points for QBench2, RealWorldQA, and MMMU, and multi-image instruction data brings 8\% point improvement on Mantis-Eval.
Our experimental results indicate that programmatically generated instruction data can indeed enhance model performance, but data scale and format are critical factors in achieving optimal results. Moreover, this programmatic instruction data proves effective for both pre-training and instruction tuning stages, and incorporating the data in both stages yields better performance than using it in either stage alone.

%% file: sec/2_method.tex
\begin{figure*}[h!]
    \centering
    \includegraphics[width=\linewidth,page=1]{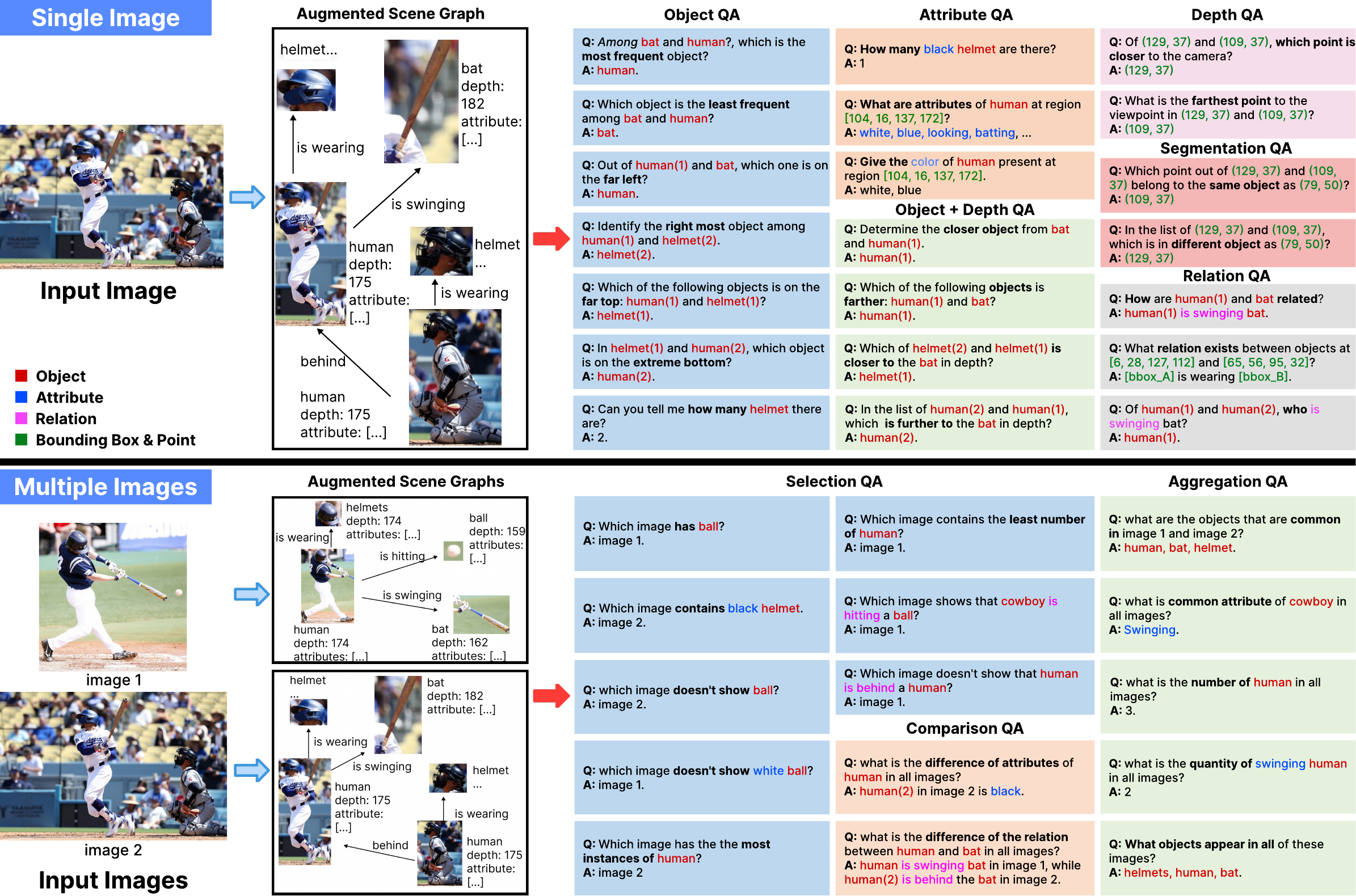}
    \caption{\label{fig:generation}We visualize the instruction data generation process and generated examples for both single and multiple image scenarios. For single image, instruction data can be categorized into six dimensions, aiming to improve the model's ability in retrieving and understanding the annotations. We divide instruction data into three categories for multiple images to help the model understand the relation between different images from the perspective of the relation between their scene graph.}
\end{figure*}

% \section{Programmatically scaling vision-centric instruction data}
\section{\method}

In this section, we first introduce how we generate vision-centric instruction data programmatically with scene graphs. Then, we present our scene graph generation pipeline that automatically generates a scene graph for any given image.

\subsection{Generating instructions programmatically}

\noindent\textbf{Augmented scene graph.}
We first describe the scene graph definition used throughout this work, which is an augmented version of the standard scene graph representation defined in Visual Genome~\cite{vg}, including additional depth and segmentation labels.
Given an input image $x$ with size $(w, h)$, which have $N$ objects $\{i_1, \cdots, i_N\}$ and each object $i_j$ has a list of attribute $a_{\text{attr}}^j$. The augmented scene graph is $G=(V, E)$, where $V\subseteq\{i_1, \cdots, i_N\}$, $E=\{(i_j, i_k, a_{\text{rel}}^{jk}) \mid i_j, i_k \in V\}$ and $a_{\text{rel}}^{jk}$ is the relation between objects $i_j$ and $i_k$. Each object $i_j$ has its corresponding bounding box and label pair $a^j_{\text{det}}$, segmentation $a^j_{\text{seg}}$, and a list of attribute $a^j_{\text{attr}}$. Additionally, we add depth annotation $a^j_{\text{dep}}$ as an augmented feature.

\noindent\textbf{Single-image visual instruction data.}
We implement 24 single-image instruction data generators to transform an augmented scene graph into thousands of high-level perceptual question-answer pairs for each image. Each generator utilizes hundreds of pre-defined templates, which systematically integrate these annotations to produce diverse instruction data. These generators are crafted to cover the model’s ability to compare, retrieve, and reason about basic visual concepts of objects, attributes, and relations based on the detailed information encoded in each scene graph.

\noindent\textbf{Multi-image visual instruction data.}
Beyond the singe-image scenarios, we also conduct 14 instruction data generators for multi-image scenarios. 
While single-image generators focus on producing instruction data from individual scene graphs, multi-image generators are capable of taking multiple scene graphs as input to generate question-answer pairs that span across images. These multi-image generators enable more complex queries, such as selection (e.g., "Which image contains more red objects?"), comparison (e.g., "What are the objects common in these images?"), and aggregation  (e.g., "How many red objects in total in these images?") questions. By leveraging multiple scene graphs simultaneously, these generators produce instruction data that encourages models to develop advanced cross-image reasoning skills.
We visualize some example instruction data that our data generators can produce in Fig.~\ref{fig:generation}.

Note that while we provide a diverse set of instruction data generators in this work, our system is designed to be highly versatile and readily extendable. It allows users to program additional data generators, expanding the space of instruction data that can be generated. By adding new templates or customizing existing ones, users can introduce novel question-answer pairs, explore new types of visual reasoning tasks, and adapt the system to meet evolving needs. This flexibility makes our framework a scalable tool for creating diverse, high-quality instruction data across various multimodal applications.

\subsection{Generating scene graph for any image}
\label{sec:sg}

\begin{figure*}[h!]
    \centering
    \includegraphics[width=\linewidth,page=1]{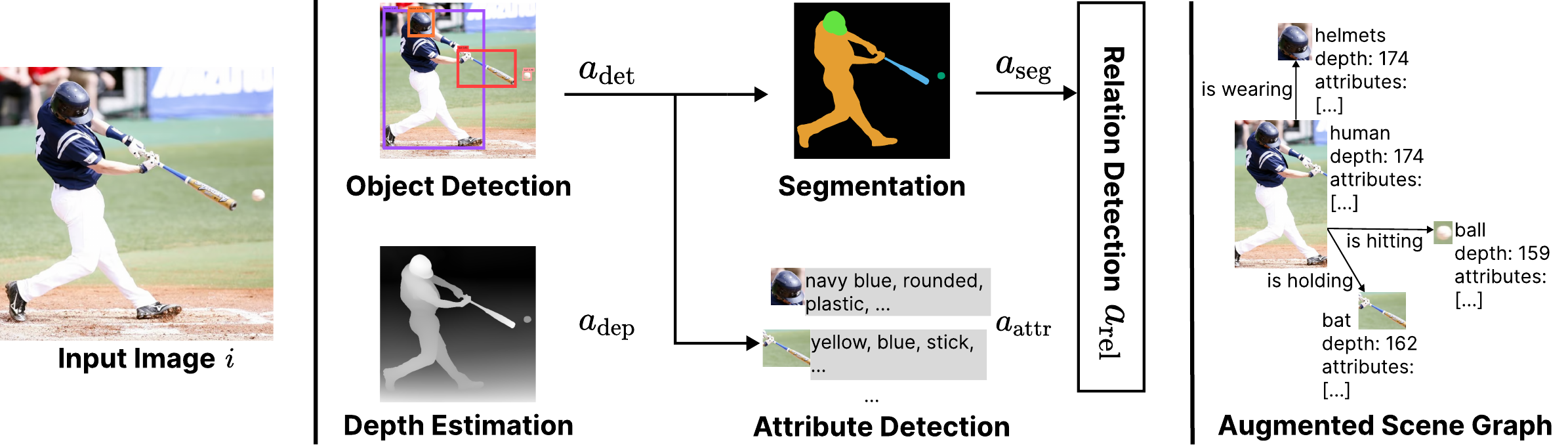}
    \caption{\label{fig:annotation}Our scene graph generation pipeline. For each input image, it will generate five annotations, including object, depth, segmentation, attribute, and relation, collectively forming an augmented scene graph.}
\end{figure*}

We generate a scene graph with object detection, image segmentation, attribute generation, relation generation, and depth estimation. While we utilize state-of-the-art, openly accessible models for each module, our approach is not limited to these specific models. We provide an overview of the scene graph generation pipeline in Fig.~\ref{fig:annotation}.

\noindent\textbf{Object detection.}
We start with object detection to seek the bounding boxes and labels for further annotation methods. The object detection model $f_{\text{det}}(x)$ will annotate all bounding boxes and the corresponding labels of all objects. For example, for object $j$, the object detection method will output $a^j_{\text{det}}=([x^j_{\text{min}}, y^j_{\text{min}}, x^j_{\text{max}}, y^j_{\text{max}}], l_j$), where $(x^j_{\text{min}}, y^j_{\text{min}})$ denotes the left bottom point of the bounding box and $l_j$ denotes the label for object $j$. In this work, we adopt YOLO-world~\cite{cheng2024yolo} as our object detection model $f_{\text{det}}(x)$.

\noindent\textbf{Image segmentation.}
We then adopt image segmentation for better object representations. The image segmentation model $f_{\text{seg}}(x, a_{\text{det}})$, SAM-2~\cite{sam2} in this work, takes the image $x$ and bounding boxes $a_{\text{det}}$ from object detection as input. Specifically, the segmentation will draw the pixel-wise segmentation $a_{\text{seg}}\in \mathbb{R}^{w\times h}$ according to $a_{\text{det}}$.

\noindent\textbf{Attribute detection.}
Inspired by prior work~\cite{zhu2024artvlm}, we finetune vision-language models, \ie, CoCa~\cite{Yu2022CoCaCC} and LLaVA-1.5~\cite{liu2024visual} as attribute detection models.
We construct the training data from LSA, a large-scale attribute dataset~\cite{Pham_2022_ECCV}. We use its bounding box annotations to crop each object as a single image and use the corresponding attribute annotation as the target output. For LLaVA-1.5, we use "$<$image$>$ \{object\_label\}" as the prompt template for finetuning data construction. Based on our automatic and manual evaluations, LLaVA-1.5-13B is better than competitors with a precision of 90\%, so we adopt it as our attribute detection model. More details of the evaluation and implementation can be found in Appendix.

\noindent\textbf{Relation Detection.}
We finally retrieve the relations $a_{\text{rel}}^{jk}$ for all pairs of objects $i_j$ and $i_k$ in image $x$ according to their segmentation. To achieve this, we pick an finetuned Osprey model~\cite{Robin} as $f_{\text{rel}}(x, a^j_{\text{seg}}, a^k_{\text{seg}})$, which takes the whole image and segmentation of objects $i_j$ and $i_k$ as input, and generate a relation $\tilde{a}_{\text{rel}}^{jk}$. 
% Specifically, the inputs are filled into the following prompt:
% \begin{tcolorbox}[mybox, title=Relation Generation Prompt, width=\linewidth]
% $<$image token$>$
% This provides an overview of the picture.
% There are \textit{region j} $<a^j_{\text{seg}}>$ $<$position for j$>$ in the image, \textit{region k} $<a^k_{\text{seg}}>$ $<$position for k$>$ in the image. 
% What is the relationship between \textit{region j} and \textit{region k}?
% \end{tcolorbox}
We then ground the generated relation by comparing the similarity between $\tilde{a}_{\text{rel}}^{jk}$ and our relation library and select the top-1 result as the $a_{\text{rel}}^{jk}$.

\noindent\textbf{Depth estimation.} Our augmented scene graph also included the pixel-wise depth annotation $a_{\text{dep}}$ generated by a depth estimator $f_{\text{dep}}(x)$. In this work, we use Depth Anything V2~\cite{yang2024depth} as our depth estimation model.
The pixel-wise depth annotation can be used to infer the depth of objects for comparing depth among objects.

%% file: sec/3_experiments.tex
\section{Experiments}

In this section, we first describe the instruction data we synthesize in this work, followed by the experimental setup, results, and analysis.
We found that 
1) Our synthesized instruction data can boost model performance and those from manually-annotated scene graphs are usually better than their counterpart from model-generated scene graphs;
2) The data format (short answer vs. multiple choice) and data scale are important factors to consider for the best performance;
and 3) While our data helps when incorporated in either the pre-training or fine-tuning stage, incorporating them in both stages achieves the best performance.
% \ranjay{This is a good place to highlight the main results so that the reader doesn't have to search for it.}

\input{tables/single_image_llava_overall_performance}

\subsection{\data Dataset Construction}

\noindent\textbf{Leveraging manually-annotated scene graph dataset.}
We first utilize Visual Genome~\cite{vg}, one large-scale manually-annotated scene graph dataset to construct our instruction data. Specifically, we augment each scene graph with depth and segmentation annotation using Depth Anything V2 and SAM-2. Then, we generate 1.5 million single-image instruction data (\vgs) and 4.2 million multi-image instruction data (\vgm):
For \vgs, we sample one instruction data per image and generator, while for \vgm, we generate 100,000 samples per generator.

\noindent\textbf{Leveraging generated scene graph.}
Besides, we sample 120,000 high-resolution images 
with more than 5 objects from the DataComp dataset~\cite{datacomp}, and use our scene graph generation pipeline as described in Sec.~\ref{sec:sg} to generate the augmented scene graph for each image.
Based on these generated scene graphs, we follow the same process as above to generate 2.3 million single-image instruction data (\dcs) and 4.2 million multi-image instruction data (\dcm).

In total, these four splits add up to more than 10 million unique instruction data to form \data. For each generated instruction, we store both a multiple-choice and a short-answer version, ensuring flexibility in the types of questions available for model training.

\subsection{Experimental Setup}

\noindent\textbf{Augmentation vs. replacement.}
To evaluate the utility of our generated dataset, we adopt two settings: augmentation and replacement.
Given a \emph{base dataset} which is an existing dataset used to train MLMs, the augmentation means augmenting the base dataset with our data, while the replacement is to replace a random subset of the base dataset with our data. In particular, we experiment with different augmentation/replacement \emph{ratios}. For example, assume the base data contains 100K samples, an augmentation ratio of 5\% indicates including an additional 5K of our data in the training set, while a replacement ratio of 5\% means replacing 5K samples of the base data with our data.

\noindent\textbf{Multiple choice vs. short answer.}
We explore both multiple choice and short answer formats for our generated instruction data, testing three distinct configurations: (1) all data in multiple choice format (multiple choice), (2) all data in short answer format (short answer), and (3) a balanced mix of formats, with half of the data in multiple choice and half in short answer (half-half). These settings allow us to assess the impact of each answer type on model performance, as well as the versatility of the generated data in supporting different response styles.

\noindent\textbf{Model and training recipe.}
We use LLaVA-1.5~\cite{liu2024visual} instruction data as the base dataset and its training recipe for instruction tuning LLaVA-1.5-7B model with single-image instruction data; similarity, we follow Mantis~\cite{jiang2024mantis} for LoRA~\cite{hu2022lora} instruction tuning Mantis-SigLIP-8B with multi-image instruction data and adopt Mantis-Instruct (excluding video-related subsets) as the base dataset.
In addition, we experiment with adding our data to both the pre-training and fine-tuning stages of xGen-MM-4B model~\cite{Xue2024xGenMMA}.

\noindent\textbf{Benchmarks.} We evaluate models on several popular MLM benchmarks including the following single-image benchmarks: CV-Bench (CVB)~\cite{tong2024cambrian1fullyopenvisioncentric}, SEED-Bench~\cite{li2023seed, li2023seed2}, MMBench (MMB)~\cite{liu2023mmbench}, MME~\cite{fu2024mmecomprehensiveevaluationbenchmark}, QBench2~\cite{wu2024qbench}, MMMU~\cite{yue2023mmmu}, RealWorldQA~\cite{xai2024grok}, MMStar~\cite{mmstar}, MMVet~\cite{mmvet}; and multi-image benchmarks: Mantis-Eval~\cite{jiang2024mantis} and MMT-Bench (MMT)~\cite{ying2024mmtbenchcomprehensivemultimodalbenchmark}.

\input{tables/multi_image_mantis_overall_performance}

\subsection{Instruction Tuning}

% We start with extensive experiments on adopting our generated instruction data in the instruction tuning stage of MLMs.
We exhibit our experiment results by answering the following questions: (1) do scene graphs help produce applicable instructions, and (2) do they need to be real, or can they be automatically generated?

\noindent\textbf{Do scene graphs help produce applicable instructions?}
% According to Table~\ref{tab:llava}, Table~\ref{tab:model-comparison}, Figure~\ref{fig:dcs}, and Figure~\ref{fig:dcm}, models trained on the dataset augmented or replaced by instructions generated from scene graphs consistently outperforms the one trained on original instruction tuning dataset, demonstrating the reliability of using scene graph to enhance the instruction tuning dataset. By deep into Table~\ref{tab:llava} and Table~\ref{tab:model-comparison}, we observe that the performance of models trained on \vgs and \vgm are positively related to the replacement or augmentation amount, which also reflects the benefit from scene graph generated instructions. 
This question can be answered affirmatively by Table~\ref{tab:llava} and Table~\ref{tab:model-comparison}. For single image instructions (Table~\ref{tab:llava}), we compare the model trained with the base dataset and the models trained on four dataset augmentation/replacement ratios and three data formats across eight benchmarking datasets. The results illustrate that, for replacement, (1) instruction tuning the LLaVA-1.5-7B model with \vgs data yields improvements over the base dataset (LLaVA-1.5 instruction data) in averaged performance across all settings and achieves the best performance when the replacement ratio at 20\%, and (2) on average, model performance is positively related to the amount of replaced multiple choice questions while negatively related to the replacement of short answers. For augmentation, we can see that (1) the model performance on all data formats increases with more data samples from \vgs and (2) compared with replacement, augmentation achieves better performance at the same level of data ratio.
Overall, results on single image tasks suggest that mixing original data with scene graph-generated short answer and multiple choice format instruction yields competitive results when a substantial portion of the original data is replaced.

For multi-image instruction (Table~\ref{tab:model-comparison}), we test the models on two multi-image benchmarks and six single-image benchmarks. We can observe that for the 20\% replacement ratio, the half-half format achieves the highest performance for both multi-image and single-image benchmarks, with an average score of 59.7, showing the benefit of mixing data formats. In contrast, at a 50\% replacement ratio, the model’s performance generally decreases in both benchmarks, suggesting that excessive replacement with new data may reduce the model's ability to generalize across tasks. For augmentation, multiple choice format stands out with a score of 60.0 on average at 20\% of augmentation, while half-half at 50\% augmentation achieves the highest average score of 60.1. This suggests that augmentation, especially with mixed data formats, can effectively enhance the model's robustness. Interestingly, augmentation generally provides higher performance stability across both multi-image and single-image benchmarks compared to replacement. On the other hand, half-half format with 10\% augmentation and multiple choice format with 20\% augmentation show strong performance across multi-image benchmarks (Mantis-Eval and MMT), showing the superiority of scene graph-generated instruction in helping the model learn how to select, compare, and aggregate features on different images.

\begin{figure}[h]
    \centering
    \includegraphics[width=0.9\columnwidth]{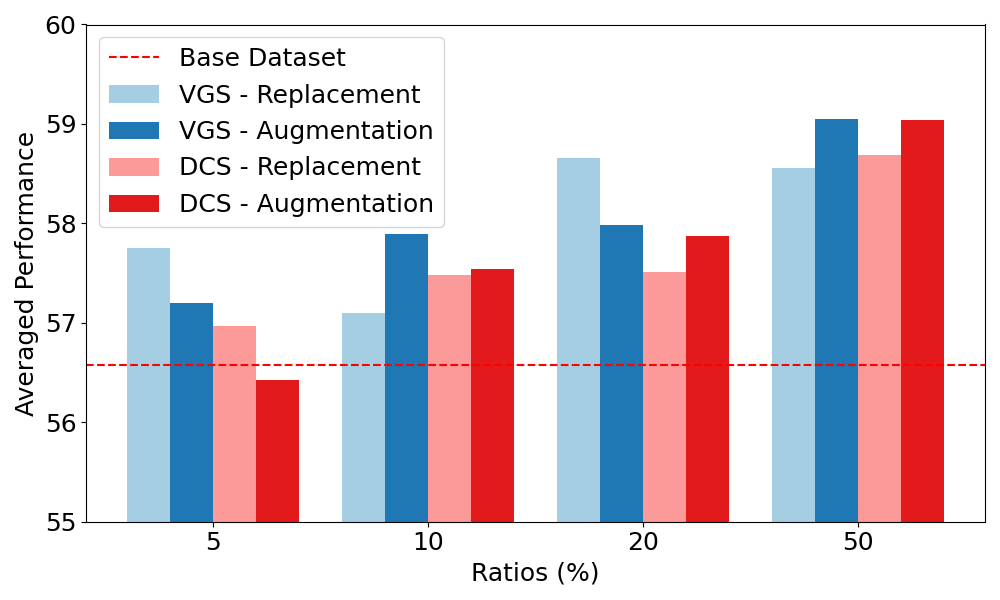}
    \caption{Results of instruction tuning LLaVA-1.5-7B with \dcs, \ie, single-image instruction data generated from the DataComp images and model-generated scene graphs.}
    \label{fig:dcs}
\end{figure}

\begin{figure}[!h]
    \centering
    \includegraphics[width=0.9\columnwidth]{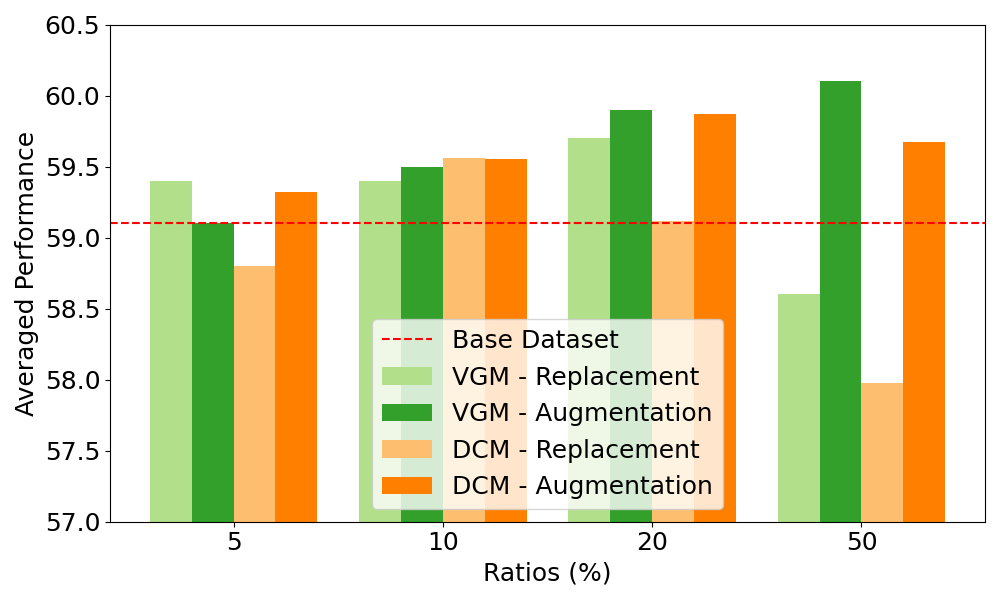}
    \caption{Results of instruction tuning Mantis-SigLIP-8B with \dcm, \ie, multi-image instruction data generated from the DataComp images and model-generated scene graphs.}
    \label{fig:dcm}
\end{figure}

\noindent\textbf{Do the scene graphs need to be manually annotated, or can they be model generated?}
We compare models trained on instruction data from manually annotated (\vgs, \vgm) and model-generated (\dcs, \dcm) scene graphs. According to Figure~\ref{fig:dcs} (\vgs vs. \dcs), \dcs underperforms \vgs at lower data scales. Interestingly, as the data scale increases to a 50\% ratio, \dcs achieves comparable performance to \vgs, suggesting that larger data scales help mitigate initial performance gaps between data from model-generated and human-curated scene graphs. 
Moreover, from Figure~\ref{fig:dcm} (\vgm vs. \dcm), we observe that as the ratio increases, the model performance on the replacement setup grows first and decays later. With the increase of replacement or augmentation ratio, \dcm underperforms with \vgm, suggesting that on multi-image settings, a larger scale on generated scene graphs may trigger edge effects and not always provide stable performance gain to the model training. 
In conclusion, instruction data from manually annotated scene graphs is in general better than that from model-generated scene graph, yet both data are able to boost model performance in most cases.

\subsection{Pre-training vs. Instruction Tuning}

\input{tables/xgen-mm_pretrain_sft_performance.tex}

To assess the benefits of incorporating our data at scale during the pre-training stage, and to compare the effects of adding our data in the pre-training versus fine-tuning stages, we adopt the xGen-MM (BLIP-3)~\cite{Xue2024xGenMMA} training methodology and use its pre-training data recipe as a foundation. We establish a baseline by pre-training a model on approximately 10 billion tokens using the xGen-MM (BLIP-3)~\cite{Xue2024xGenMMA}  pre-training recipe without our data. Similarly, we apply a baseline fine-tuning recipe of 1 million samples that excludes our data. Details for both recipes are provided in the Appendix.

In the augmentation setup mentioned previously, we create recipes that incorporate our data into both the pre-training and fine-tuning stages. \textbf{At each stage, we ensure that our additional data accounts for around 5\% of the final dataset size}. To examine the impact of different types of scene graph data input on our single-image instruction data generation pipeline, we conduct two sets of experiments: one using human-annotated scene graphs (\vgs) as the data source, and another using synthesized scene graphs generated by our pipeline (\dcs). Results in table~\ref{tab:pretrain-sft-comparison} reveal several insights:

\begin{enumerate}
    \item  \textbf{Performance Gains with Augmentating Our Data in the Base Recipes}: Both \dcs and \vgs augmentations consistently improve model performance across the 11 evaluated benchmarks compared to the baseline dataset, which does not include our data. This validates the efficacy of incorporating additional vision-centric knowledge into multimodal foundation model training.
    \item  \textbf{Synergistic Effect of Dual-Stage Augmentation}: The results suggest that augmentation during both pre-training and fine-tuning stages synergistically enhances the performance. For instance, models trained on datasets augmented in both stages achieve the highest average scores (e.g., \dcs with +1.2\% and \vgs with +1.6\%), indicating that the synergized effect of dual-stage augmentation outperforms augmentation at either stage alone.
    \item \textbf{Comparative Effectiveness of \vgs and \dcs}: While both types of data yield improvements, \vgs-based augmentation generally provides a marginally higher average score (60.1\%) compared to \dcs (59.7\%) when augmented in both stages. This suggests that cleaner scene graph data source may enable our pipeline to yield even higher performance.
\end{enumerate}

% when augmented with both the \dcs or \vgs, the model performances averaged on 12 benchmarks all beat the baseline model's performance (without adding any of our data). Also we have a few observations that: 1. when augment in either pre-train stage and fine-tune stage, the performance will improve, and when augment in both pre-train and fine-tune stages, the benefits seem to synergize and improve more. 

% One advantage of our pipeline is its scalability, which enables our pipeline to be applied in large scale multimodal pre-training. We conduct experiments comparing the addition of 0.75 million versus 1.5 million samples of \vgs during pre-training, while keeping the fine-tuning recipe consistent. As shown in table ~\ref{tab:scaling-comparison}, When scaling from 0.75 million to 1.5 million \textbackslash{}vgs samples, we observe notable gains in the average performance, from 61.4\% to 62.3\%, underscoring the potential of our data in large scale multimodal training.
A key advantage of our pipeline is its scalability, enabling its application in large-scale multimodal language model training. We conduct experiments comparing the addition of 0.75 million versus 1.5 million samples of \vgs during pre-training, while keeping the fine-tuning recipe consistent. As shown in Table~\ref{tab:scaling-comparison}, scaling the inclusion of \vgs in pre-training stage from 0.75 million to 1.5 million samples yields notable gains in average performance across 12 benchmarks, increasing from 61.4\% to 62.3\%. This result underscores the potential of our data to enhance model capabilities in large-scale multimodal foundation model training.

%% file: tables/single_image_llava_overall_performance.tex
\begin{table*}[!t]
  \centering
  \small
  \scalebox{0.80}{
    \begin{tabular}{cccccccccccc} 
    \toprule
    & \bf Data Ratio* & \bf Data Format & \bf CVB-2D & \bf CVB-3D & \bf SEED & \bf MMB & \bf MME & \bf QBench2 & \bf MMMU & \bf RealWorldQA & \bf Avg. \\
    \midrule
    \multicolumn{3}{c}{LLaVA-1.5 instruction data~\cite{liu2024visual}} & 58.0 & 61.0 & 66.8 & 66.7 & 63.2 & 46.4 & 36.2 & 54.2 & 56.6 \\ 
\midrule
\midrule
\multirow{12}{*}{\bf Replacement} & \multirow{3}{*}{5\%} 
& Short Answer & 55.0 & 66.0 & 67.1 & 66.3 & 64.2 & 48.5 & 36.7 & 52.7 & 57.1 \\ 
& & Multiple Choice & 61.0 & 61.0 & \textbf{67.5} & 67.0 & 63.3 & 47.8 & 37.8 & 54.6 & 57.5 \\ 
& & Half-Half & 60.0 & 66.0 & 67.4 & 66.5 & 62.5 & 46.6 & 37.4 & 55.6 & 57.8 \\  
\cmidrule{2-12}
& \multirow{3}{*}{10\%} 
& Short Answer & 58.0 & 67.0 & 67.0 & 67.4 & 64.2 & 49.2 & 37.8 & 56.1 & 58.3 \\ 
& & Multiple Choice & 56.0 & 62.0 & 67.2 & 67.1 & \textbf{64.4} & 48.4 & 36.8 & \textbf{58.7} & 57.6 \\ 
& & Half-Half & 56.0 & 64.0 & 67.0 & 67.3 & 63.4 & 46.5 & 36.7 & 56.0 & 57.1 \\  
\cmidrule{2-12}
& \multirow{3}{*}{20\%} 
& Short Answer & 59.0 & 66.0 & 67.3 & 66.8 & 63.4 & 47.7 & 36.1 & 54.5 & 57.6 \\ 
& & Multiple Choice & 57.0 & 63.0 & 66.8 & \textbf{68.0} & 63.2 & 48.9 & 37.2 & 58.6 & 57.8 \\
& & Half-Half & 63.0 & 66.0 & 67.5 & 66.7 & 62.5 & 46.7 & \textbf{39.1} & 57.9 & \textbf{58.7} \\
\cmidrule{2-12}
& \multirow{3}{*}{50\%} 
& Short Answer & 54.0 & \textbf{69.0} & 65.9 & 64.9 & 60.5 & \textbf{50.2} & 35.2 & 55.7 & 56.9 \\
& & Multiple Choice & 61.0 & 68.0 & 66.3 & 66.1 & 61.8 & 46.1 & 38.2 & 56.7 & 58.0 \\
& & Half-Half & \textbf{65.0} & \textbf{69.0} & 66.6 & 65.0 & 62.3 & 47.2 & 38.1 & 55.3 & 58.6 \\ 
\midrule 
\midrule
\multirow{12}{*}{\bf Augmentation} & \multirow{3}{*}{5\%} 
& Short Answer & 55.0 & 65.0 & 66.7 & 66.5 & 63.9 & 48.1 & 37.3 & 55.6 & 57.2 \\ 
&& Multiple Choice & 60.0 & 63.0 & 66.5 & 67.7 & 64.1 & 47.7 & 37.8 & 55.7 & 57.8 \\ 
&& Half-Half & 56.0 & 66.0 & 66.8 & 67.1 & 61.6 & 46.5 & 37.6 & 56.1 & 57.2 \\  
\cmidrule{2-12}
&\multirow{3}{*}{10\%} 
& Short Answer & 59.0 & \textbf{69.0} & 66.7 & 66.9 & 62.3 & 49.0 & 37.3 & 54.0 & 58.0 \\
&& Multiple Choice & 60.0 & 64.0 & 66.4 & 66.7 & 63.3 & 47.6 & 37.1 & 56.0 & 57.6 \\
&& Half-Half & 57.0 & 67.0 & 68.0 & \textbf{68.1} & 64.2 & 45.3 & 38.4 & 55.0 & 57.9 \\
\cmidrule{2-12}
&\multirow{3}{*}{20\%} 
& Short Answer & 59.0 & 68.0 & 67.2 & 68.0 & 61.6 & 48.5 & 37.6 & 54.0 & 58.0 \\ 
&& Multiple Choice & 58.0 & 63.0 & 67.0 & 67.7 & 63.3 & 46.2 & 37.6 & 56.6 & 57.4 \\
&& Half-Half & 58.0 & 67.0 & 67.7 & 67.2 & 63.0 & 46.7 & 36.4 & 57.9 & 58.0 \\ 
\cmidrule{2-12}
&\multirow{3}{*}{50\%} 
& Short Answer & 57.0 & \textbf{69.0} & 67.2 & 68.0 & 64.5 & \textbf{49.4} & 37.2 & 55.3 & 58.5 \\
&& Multiple Choice & 61.0 & 68.0 & 67.4 & 67.8 & 63.3 & 48.5 & 36.3 & 56.6 & 58.6 \\ 
&& Half-Half & 60.0 & 66.0 & 67.5 & 67.6 & \textbf{66.0} & 48.7 & \textbf{38.9} & 57.6 & \textbf{59.0} \\
\midrule

    \end{tabular}
  }
    \caption{Results of instruction tuning LLaVA-1.5-7B with \vgs, \ie, single-image instruction data generated from the Visual Genome images and scene graphs. *: data ratio = number of our data added  / size of LLaVA-1.5 instruction data.}
  \label{tab:llava}
\end{table*}

%% file: tables/multi_image_mantis_overall_performance.tex
\begin{table*}[!t]
  \centering
  \small

  \scalebox{0.80}{
    \begin{tabular}{c c c c c c c c c c c c} 
    \toprule
    \multicolumn{3}{c}{} &  \multicolumn{2}{c}{\bf Multi-image benchmark} &   \multicolumn{6}{c}{\bf Single-image benchmark} \\
    \cmidrule(lr){4-5}\cmidrule(lr){6-11}
    & \bf Data Ratio* & \bf Data Format & \bf Mantis-Eval & \bf MMT & \bf SEED & \bf MMB & \bf MME & \bf QBench2 & \bf MMMU & \bf RealWorldQA & \bf Avg. \\
    \midrule
    \multicolumn{3}{c}{Mantis instruction data~\cite{jiang2024mantis}} & 54.4 & 52.9 & 68.1 & 72.8 & 58.5 & 70.1 & 44.3 & 51.5 & 59.1 \\
\midrule
\midrule
\multirow{12}{*}{\bf Replacement} & \multirow{3}{*}{5\%} 
& Short Answer & 59.9 & 52.5 & 68.4 & 73.3 & \textbf{60.0} & 70.3 & 41.4 & 50.3 & 59.5 \\ 
& & Multiple Choice & 57.6 & 52.8 & 68.0 & 73.0 & 58.4 & 70.5 & 43.7 & 52.3 & 59.5 \\
& & Half-Half & 57.6 & 53.4 & 68.1 & 71.7 & 57.8 & \textbf{71.5} & 44.4 & 50.9 & 59.4 \\  
\cmidrule{2-12}
& \multirow{3}{*}{10\%} 
& Short Answer & 69.0  & \textbf{54.8} & 68.6 & \textbf{73.7} & 57.4 & 68.4 & 41.3 & 50.7 & 59.2 \\ 
& & Multiple Choice & 59.9 & 53.1 & 68.3 & 71.7 & 57.5 & 68.6 & 45.3 & 51.9 & 59.5 \\ 
& & Half-Half & 59.0 & 53.7 & 68.2 & 72.6 & 58.6 & 68.9 & 43.1 & 51.5 & 59.4 \\ 
\cmidrule{2-12}
& \multirow{3}{*}{20\%} 
& Short Answer & \textbf{62.7} & 52.9 & 68.2 & 72.9 & 56.6 & 70.2 & \textbf{45.4} & 49.7 & \textbf{59.8} \\ 
& & Multiple Choice & 57.6 & 52.2 & 68.0 & 72.9 & 57.7 & 67.4 & 42.0 & 51.4 & 58.6 \\
& & Half-Half & 58.5 & 58.6 & \textbf{68.7} & 72.2 & 59.6 & 69.4 & 44.1 & 51.8 & 59.7 \\
\cmidrule{2-12}
& \multirow{3}{*}{50\%} 
& Short Answer & 57.1 & 53.2 & 67.4 & 70.4 & 58.6 & 65.2 & 42.2 & 51.2 & 58.2 \\
& & Multiple Choice & 55.8 & 52.5 & 67.5 & 69.8 & 57.5 & 67.9 & 42.6 & \textbf{53.5} & 58.4 \\
& & Half-Half & 54.8 & 54.0 & 67.9 & 72.1 & 58.2 & 66.8 & 43.7 & 51.5 & 58.6 \\
\midrule 
\midrule
\multirow{12}{*}{\bf Augmentation} & \multirow{3}{*}{5\%} 
& Short Answer & 60.4 & 53.8 & 68.3 & 71.2 & 58.9 & 70.6 & \textbf{44.3} & 48.9 & 59.5 \\
&& Multiple Choice & 58.1 & 54.0 & 68.1 & 71.7 & 58.8 & 70.3 & 42.4 & 50.2 & 59.2 \\ 
&& Half-Half & 58.1 & 52.5 & 68.0 & 71.8 & 58.4 & 70.1 & 41.3 & \textbf{52.9} & 59.1 \\ 
\cmidrule{2-12}
&\multirow{3}{*}{10\%} 
& Short Answer & 60.4 & 53.0 & 68.1 & 72.8 & 59.2 & 71.4 & 44.2 & 50.6 & 60.0 \\
&& Multiple Choice & 60.4 & 52.7 & 68.0 & 72.2 & 59.2 & 71.1 & 43.1 & 50.2 & 59.6 \\
&& Half-Half & \textbf{61.3} & 53.0 & 68.0 & 72.9 & 60.0 & 67.7 & 42.0 & 51.0 & 59.5 \\ 
\cmidrule{2-12}
&\multirow{3}{*}{20\%} 
& Short Answer & 57.1 & 53.2 & 68.5 & 72.3 & 60.3 & \textbf{71.6} & 43.8 & 50.3 & 59.6 \\ 
&& Multiple Choice & 58.5 & 52.9 & \textbf{68.6} & 72.1 & \textbf{60.6} & \textbf{71.6} & 43.0 & 51.5 & 60.0 \\
&& Half-Half & 60.4 & 52.8 & 68.5 & 72.4 & \textbf{60.6} & 68.4 & 43.7 & 52.7 & 59.9 \\ 
\cmidrule{2-12}
&\multirow{3}{*}{50\%} 
& Short Answer & 57.6 & 53.0 & 68.1 & 71.6 & 59.1 & 70.4 & \textbf{44.3} & 51.2 & 59.4 \\
&& Multiple Choice & 58.5 & \textbf{54.1} & 68.4 & 72.4 & 58.8 & 70.1 & 41.4 & 51.6 & 59.4 \\ 
&& Half-Half & 60.4 & 53.7 & 68.1 & \textbf{73.4} & 60.4 & 69.7 & 43.8 & 51.6 & \textbf{60.1} \\
\midrule

    \end{tabular}
  }
    \caption{Results of instruction tuning Mantis-SigLIP-8B with \vgm, \ie, multi-image instruction data generated from the Visual Genome images and scene graphs. *: data ratio = number of our data added  / size of Mantis instruction data.}
  \label{tab:model-comparison}
\end{table*}

%% file: tables/xgen-mm_pretrain_sft_performance.tex
% \begin{table*}[!h]
%   \centering
%   \small
%   \caption{xGen-MM (BLIP-3) Performance Comparison}
%   \scalebox{0.75}{
%     \begin{tabular}{c|c|c|c|c|c|c|c|c|c|c|c|c} 
%     \toprule
%     \bf Training Data & \bf CVB-2D & \bf CVB-3D & \bf SEED & \bf MMB & \bf MMStar & \bf MME & \bf MMVet & \bf MMMU & \bf RealWorldQA & \bf POPE & \bf TextVQA & \bf Avg. \\
%     \midrule
%     Pre-train wo/ + SFT wo/ (baseline) & 62.9 & 73.5 & 70.1 & 73.9 & 44.1 & 62.1 & 38.2 & 41.6 & 56.2 & 87.3 & 66.5 & 61.5 \\ 
%     \midrule
%     \multicolumn{13}{c}{\bf Images and Scene Graphs from VG} \\ 
%     \midrule
%     Pre-train wo/ + SFT w/ & 67.7 & 72.2 & 70.0 & 73.6 & 45.5 & 64.2 & 36.7 & 42.9 & 60.4 & 86.6 & 67.3 & 62.5 \\ 
%     Pre-train w/ + SFT wo/ & 65.9 & 73.3 & 70.5 & 75.9 & 44.8 & 64.9 & 36.8 & 42.2 & 59.2 & 86.2 & 67.3 & 62.5 \\ 
%     Pre-train w/ + SFT w/ & 70.2 & 73.7 & 69.9 & 73.9 & 44.3 & 64.0 & 40.4 & 44.8 & 57.4 & 86.7 & 66.8 & 62.9 \\ 
%     \midrule
%     \multicolumn{13}{c}{\bf Images from Datacomp-1B and Scene Graphs are Synthesized from Our Pipeline} \\ 
%     \midrule
%     Pre-train wo/ + SFT w/ & 64.5 & 71.9 & 69.1 & 73.3 & 44.8 & 60.9 & 35.2 & 44.1 & 59.7 & 86.7 & 67.0 & 61.6 \\ 
%     Pre-train w/ + SFT wo/ & 67.9 & 70.8 & 70.4 & 73.5 & 45.5 & 64.4 & 37.7 & 46.1 & 58.8 & 86.3 & 67.1 & 62.6 \\ 
%     Pre-train w/ + SFT w/ & 68.3 & 75.5 & 69.6 & 74.5 & 44.4 & 62.4 & 38.5 & 41.9 & 61.6 & 86.4 & 66.5 & 62.7 \\ 
%     \midrule
%     \end{tabular}
%   }
%   \label{tab:model-comparison}
% \end{table*}

\begin{table*}[!btp]
  \centering
  \small
  
  \scalebox{0.73}{
    \begin{tabular}{c|c|c|c|c|c|c|c|c|c|c|c|c|c} 
    \toprule
    \bf Augment in & \bf Augment in & \multirow{2}{*}{\bf CVB-2D} & \multirow{2}{*}{\bf CVB-3D} & \multirow{2}{*}{\bf SEED} & \multirow{2}{*}{\bf MMB} & \multirow{2}{*}{\bf MMStar} & \multirow{2}{*}{\bf MME} & \multirow{2}{*}{\bf QBench2} & \multirow{2}{*}{\bf MMVet} & \multirow{2}{*}{\bf MMMU} & \multirow{2}{*}{\bf RealWorldQA} & \multirow{2}{*}{\bf TextVQA} & \multirow{2}{*}{\bf Avg.} \\
    \bf Pre-train & \bf Fine-tune & & & & & & & & & & & & \\
    \midrule
    \xmark & \xmark & 62.9 & 73.5 & 70.1 & 73.9 & 44.1 & 62.1 & 53.9 & 38.2 & 41.6 & 56.2 & 66.5 & 58.5 \\ 
    \midrule
    \multicolumn{14}{c}{\bf \dcs:  DataComp images and model-generated synthetic scene graphs} \\ 
    \midrule
    \xmark & \cmark & 64.5 & 71.9 & 69.1 & 73.3 & 44.8 & 60.9 & \textbf{57.9} & 35.2 & 44.1 & 59.7 & 67.0 & 58.9 \\ 
    \cmark & \xmark & 67.9 & 70.8 & \textbf{70.4} & 73.5 & \textbf{45.5} & \textbf{64.4} & 53.6 & 37.7 & \textbf{46.1} & 58.8 & \textbf{67.1} & 59.6 \\ 
    \cmark & \cmark &\textbf{ 68.3} & \textbf{75.5} & 69.6 & \textbf{74.5} & 44.4 & 62.4 & 54.0 & \textbf{38.5} & 41.9 & \textbf{61.6} & 66.5 & \textbf{59.7} \\ 
    \midrule
    \multicolumn{14}{c}{\bf \vgs: Visual Genome images and scene graphs} \\ 
    \midrule
    \xmark & \cmark & 67.7 & 72.2 & 70.0 & 73.6 & \textbf{45.5} & 64.2 & 54.4 & 36.7 & 42.9 & \textbf{60.4} & \textbf{67.3} & 59.5 \\ 
    \cmark & \xmark & 65.9 & 73.3 & \textbf{70.5} & \textbf{75.9} & 44.8 & \textbf{64.9} & \textbf{56.3} & 36.8 & 42.2 & 59.2 & \textbf{67.3} & 59.7 \\ 
    \cmark & \cmark & \textbf{70.2} & \textbf{73.7} & 69.9 & 73.9 & 44.3 & 64.0 & 55.7 & \textbf{40.4} & \textbf{44.8} & 57.4 & 66.8 & \textbf{60.1} \\ 
    \midrule
    \end{tabular}
    }
    \caption{Comparison of augmenting the base data recipe with our data (both \dcs and \vgs) in pre-training, fine-tuning, or both. The first row represents the baseline model's performance, which uses only the base recipes for both stages without our data. The results demonstrate that augmenting the base data recipes with our data in either stage (pre-training or fine-tuning) improves the model’s average performance across 11 benchmarks, with augmentation in both stages generally achieving the highest average performance. When augmented with 1.5 million samples from \vgs, our model shows an average improvement of approximately 1.4\% over the baseline across all benchmarks. Bold values indicate the highest score for each benchmark.}

  \label{tab:pretrain-sft-comparison}
\end{table*}

\begin{table}[!btp]
  \centering
  \small
  % \caption{Scaling Experiments}
  \scalebox{1.0}{
    \begin{tabular}{c|c} 
    \toprule
    \bf Dataset Augmentation Scale & \bf Avg. \\
    \midrule
    No Augmentation & 58.5 \\ 
    0.75 Million & 59.1\\ 
    1.5 Million & \bf 60.1 \\
    \midrule
    \end{tabular}
  }
  \caption{Impact of dataset augmentation scale on model performance. Under the setup of augmenting data in both pre-training and fine-tuning stages, this table compares the model's performance across the same 11 benchmarks as Table~\ref{tab:scaling-comparison}, with varying scales of data augmentation during pre-training: no augmentation, 0.75 million samples, and 1.5 million samples from \vgs.
  % The results show that increasing the augmented scale of our data generally enhances average performance.
  % , with 1.5 million samples achieving the highest average score of 62.3\%. 
  % Bold values indicate the highest score for each benchmark.
  }
  \label{tab:scaling-comparison}
  \vspace{-5mm}
\end{table}

% \begin{table*}[!btp]
%   \centering
%   \small
%   % \caption{Scaling Experiments}
%   \scalebox{0.74}{
%     \begin{tabular}{c|c|c|c|c|c|c|c|c|c|c|c|c|c} 
%     \toprule
%     \bf Data Augmentation Scale & \bf CVB-2D & \bf CVB-3D & \bf SEED & \bf MMB & \bf MMStar & \bf MME & \bf QBench2 & \bf MMVet & \bf MMMU & \bf RealWorldQA & \bf POPE & \bf TextVQA & \bf Avg. \\
%     \midrule
%     No Augmentation & 62.9 & 73.5 & 70.1 & 73.9 & 44.1 & 62.1 & 53.9 & 38.2 & 41.6 & 56.2 & \textbf{87.3} & 66.5 & 60.9 \\ 
%     0.75 Million & 66.7 & 71.1 & \textbf{70.3} & \textbf{74.2} & \textbf{44.6} & 62.4 & 52.7 & 38.7 & 44.2 & \textbf{58.3} & 86.9 & \textbf{67.0} & 61.4\\ 
%     1.5 Million & \textbf{70.2} & \textbf{73.7} & 69.9 & 73.9 & 44.3 & \textbf{64.0} & \textbf{55.7} & \textbf{40.4} & \textbf{44.8} & 57.4 & 86.7 & 66.8 & \bf 62.3 \\
%     \midrule
%     \end{tabular}
%   }
%   \caption{Impact of data augmentation scale of our data during the pre-training stage on model performance. This table compares the model's performance across 12 benchmarks with varying scales of data augmentation during pre-training: no augmentation, 0.75 million samples, and 1.5 million samples from \vgs. The results show that increasing the augmented scale of our data generally enhances average performance.
%   % , with 1.5 million samples achieving the highest average score of 62.3\%. 
%   Bold values indicate the highest score for each benchmark.}
%   \label{tab:scaling-comparison}
% \end{table*}

%% file: sec/4_related.tex
\section{Related Work} 

We contextualize our work on the recent rise of MLMs and approaches of synthesizing data for MLMs.

\noindent\textbf{Multimodal language models (MLMs).}
In recent years, MLMs, by integrating visual encoders within various pretrained large languages models~\cite{Wang2024InternVideo2SV, huang2023vtimellm, chen2023videollm, wang2023gpt4video, sun2023finegrained, lyu2023macawllm, tang2023llmvagebc, wang2023chatvideo, lin2023mmvid, bi2023misar, chen2023groundingprompter, liu2024prismer, peng2023kosmos2, chen2023pali3, shukor2023unival, lin2023mmvid, lu2023chameleon, li2023mimicit, sun2024emu, moor2023medflamingo, awadalla2023openflamingo, sun2024generative, Xue2024xGenMMA}, have progressively driven advancements in visual-language learning. With ubiquitous open-sourced LLM backbones and the increasing data for visual instruction tuning. Models like Blip series~\cite{dai2024instructblip, li2022blip, li2023blip2, Panagopoulou2023XInstructBLIPAF, Xue2024xGenMMA}, QwenVL series~\cite{bai2023qwen, Qwen2VL}, LLaVA series~\cite{liu2024visual, liu2023improvedllava, liu2024llavanext}, InternVL series~\cite{chen2023internvl, chen2024far}, etc. have achieved unprecedented visual understanding performance in nearly all kind of visual tasks. 
% These models, which take both visual content and language as input and output languages, are being considered as a new type of foundation model with unprecedented capabilities in visual understanding. 
However, recent works like Task Me Anything~\cite{zhang2024task}, CVBench (Cambrian-1)~\cite{tong2024cambrian1fullyopenvisioncentric} show that while MLMs are adept at high-level semantic understanding, they surprisingly underperform in vision-centric tasks (e.g. depth estimation, counting, localization, etc). Furthermore, the availability of instruction data for vision-centric tasks remains limited compared to other multimodal data such as image captions, due to the high cost of collection and annotation. 
% To tackle this problem, we provide a programmatic vision-centric instruction data pipeline that can automatically generate large-scale data for MLMs without human annotation.

\noindent\textbf{Synthetic data for MLMs.}
Synthetic data has increasingly been used for pretraining and finetuning~\cite{Dong2024SelfBoostingLL, Narayan2024CookbookAF, Yu2024CodePMPSP, Chan2024BalancingCA, Zhou2024ProgrammingEE, yang2024synthetic, Zhao2024SELFGUIDEBT, Xu2024MagpieAD, Zhou2024JiuZhang30EI, Wang2024CodecLMAL, Lee2024LLM2LLMBL, Nayak2024LearningTG, liu2024mminstruct, liu2024mminstructgeneratedvisualinstructions, li2024DenseFusion,awadalla2024blip3kaleknowledgeaugmentedlargescale, Panagopoulou2023XInstructBLIPAF} of large language models (LLMs), leading to notable improvements in reasoning, instruction following, and other tasks. Similarly, synthetic data has been integrated into multimodal language model (MLM) development, including approaches like model-generated instruction data~\cite{liu2024visual,wang2023instruct4v,liu2024mminstructgeneratedvisualinstructions,liu2024mminstruct} and synthetic captions~\cite{li2024DenseFusion,Liu2024SynthVLMHA, Sharifzadeh2024Synth2BV,yan2024list}. However, current methods largely focus on synthetic data generation using LLM, MLM, and diffusion models. Programmatic/procedural methods have also been employed to generate multimodal data, such as in GQA~\cite{hudson2019gqa}, AGQA~\cite{GrundeMcLaughlin2021AGQA}, and Task Me Anything~\cite{zhang2024task}, yet these are often designed primarily for evaluation or as contributions to a final dataset. In contrast, our approach centers on the data generation process itself, producing single- and multi-image instruction data adaptable to any image source for training purposes.
% Cambrian-1~\cite{} introduces a benchmark to assess the computer vision (CV) capabilities of MLMs. Despite this progress, the availability of high-quality, CV-related instruction data tailored for MLMs remains limited. In this paper, we propose a programmatic approach to address this gap by generating synthetic data for CV tasks in a way that facilitates effective training and evaluation of MLMs.

%% file: sec/5_conclusion.tex
\section{Conclusion}

% \ranjay{Try and not repeating / summarizing the abstract but use this space to discuss limitations and future directions instead.}
Our \method system programmatically synthesizes vision-centric instruction data for training MLMs by leveraging scene graph representations and human-written programs. Applied to Visual Genome and DataComp, \method produces \data, a dataset of over 10 million instruction data, which we leverage in both pretraining and instruction tuning stages of MLMs, resulting in notable performance improvements and demonstrating the potential of programmatically scaling vision-centric instruction data in advancing MLM capabilities.

\noindent\textbf{Limitations and future directions.}
Limitations of \method include its reliance on the quality and completeness of scene graphs, as well as its dependency on human-written programs. Future work could address these by enhancing the scene graph generation pipeline to enable more accurate data synthesis and by developing automated program synthesis, leveraging LLMs to further scale data generation.

%% file: sec/X_suppl.tex
\clearpage

\setcounter{page}{1}
\maketitlesupplementary
\section{Fine-tuning attribute detector}

\paragraph{Dataset preparation.}
We adopt the LSA dataset~\cite{Pham_2022_ECCV} for training attribute detectors for our scene graph generation pipeline. We first filter out bounding boxes whose size is less than 25 pixels. Then we normalize the attributes by (1) removing non-attributes like "world" and (2) merging attributes like "gray" and "grey". We also remove objects with conflicting attributes at the same time like "big" and "small". Finally, for each object category with more than 10 instances in the dataset, we sample 5 instances to compose the test set (42,558 objects) and use the remaining as the training set (3,679,514 objects).

\paragraph{Fine-tuning and evaluation.} For CoCa model, we use the OpenCLIP codebase to fine-tune a ViT-L-14 CoCa pretrained on LAION-2B. For LLaVA-1.5 model, we use the official codebase to fine-tune both LLaVA-1.5-7B and LLaVA-1.5-13B.
For evaluation, we report the average number of predicted attributes of each model. Besides, we report the precision and recall of the model output against the provided labels. However, because the LSA dataset is noisy and incomplete, we sample 200 data from the test set and manually evaluate whether each predicted attribute is correct or not to calculate the human precision.

\begin{table*}[h]
  \centering
  \small
  % \resizebox{\linewidth}{!}{
    \begin{tabular}{ccccc} 
    \toprule
    {\bf Model} & {\bf Avg. number of predicted attributes} & {\bf Precision} & {\bf Recall} & {\bf Human precision}\\ 
    \midrule

    \multirow{1}{*}{CoCa}  
    & 3.19 & 14.5 & 36.0 & 53.1\\ 
    
    \midrule

    \multirow{1}{*}{ LLaVA-1.5-7B}  
    &  1.13 & 57.3 & 50.4 & 90.5 \\ 

\midrule
    
    \multirow{1}{*}{ LLaVA-1.5-13B}  
    & 1.13 & 58.4 & 51.5 & 91.9\\ 
    
\bottomrule 
\end{tabular}
% }

\caption{\label{tab:attr}Results of fine-tuned attribute detector.}
\end{table*}

\paragraph{Results.} 
The results are in Table~\ref{tab:attr}.
From the results, we can see that the CoCa model outputs more attributes (3.19) than LLaVA-1.5 models (1.13), because the LLaVA-1.5 models take the object label as input and is likely to focus on the object in the cropped image while CoCa only inputs the cropped image and may output attributes irrelevant to the centered object.
In addition, we found that LLaVA-1.5-13B is better than LLaVA-1.5-7B and CoCa for all the evaluation metrics, so we adopt LLaVA-1.5-13B for our scene graph generation pipeline.

We will release both train/test dataset and the trained models.

\section{Instruction Tuning Experiments}

For fine-tuning LLaVA-1.5 models, we reuse the fine-tuning script in the official Github repository: \url{https://github.com/haotian-liu/LLaVA/blob/main/scripts/v1_5/finetune.sh}.

For fine-tuning Mantis-SigLIP-8B, we reuse the script from the official Github repository: \url{https://github.com/TIGER-AI-Lab/Mantis/blob/main/mantis/train/scripts/train_mllava.sh}.

\section{xGen-MM (BLIP-3) Experiments Recipes}
\subsection{Pre-training Recipes}

\paragraph{Base Recipe.}  
Following xGen-MM(BLIP-3)~\cite{Xue2024xGenMMA}, the base pre-training recipe includes the following --  
\textbf{Caption Datasets:} Datacomp~\cite{datacomp} (10\%), BLIP3-KALE~\cite{awadalla2024blip3kaleknowledgeaugmentedlargescale} (60\%), BLIP3-OCR~\cite{Xue2024xGenMMA} (10\%), BLIP3-GROUNDING~\cite{Xue2024xGenMMA} (10\%), CC12M~\cite{cc12m} (2.5\%), CC3M~\cite{cc12m} (2.5\%), VG~\cite{vg} (2.5\%), and SBU~\cite{sbu} (2.5\%).  
\textbf{Interleaved Datasets:} OBELICS~\cite{laurenccon2024obelics} (35\%), MINT-1T-HTML~\cite{mint1t} (35\%), MINT-1T-PDF~\cite{mint1t} (25\%), and MINT-1T-ArXiv~\cite{mint1t} (5\%).  

The baseline model is trained using 24 H100-80GB GPUs for 8,000 steps. At each step, data is sampled with equal probability from either the \textbf{Caption Datasets} or the \textbf{Interleaved Datasets} bucket (50\% each). Within each bucket, datasets are sampled according to the probabilities listed above.  

The batch sizes are configured as follows:  
\textbf{Caption Datasets:} Batch size of 300 per GPU.  
\textbf{Interleaved Datasets:} Batch size of 50 per GPU.  

\paragraph{Augmented Recipe with PROVISION Data.}  
To ensure a fair comparison, the composition of the \textbf{Caption Datasets} and \textbf{Interleaved Datasets} from the base recipe is preserved. Additionally, a new dataset bucket, \textbf{PROVISION}, is introduced. The sampling ratios are adjusted from 50\%:50\% (Caption Datasets vs Interleaved Datasets) to 47.5\%:47.5\%:5\% (Caption Datasets vs Interleaved Datasets vs PROVISION).  

The training duration is extended from 8,000 steps to 8,500 steps to ensure that the amount of caption and interleaved data remains consistent while incorporating the PROVISION data. All other training configurations are kept unchanged, allowing a fair assessment of the impact of including PROVISION data.

\subsection{Fine-tuning Recipes}
We use a mixture of open-source supervised fine-tuning datasets~\cite{liu2024llavanext, idefics2, tong2024cambrian1fullyopenvisioncentric, yan2024list, li2024llava} as our base SFT data recipe. The base SFT data recipe contains around 1.2M single-image QA samples. We create the base SFT recipe to cover various visual tasks including:
\begin{itemize}
    \item \textbf{General visual question answering (630K)}: sharegpt4v~\cite{chen2023sharegpt4v}, sharegpt4o~\cite{sharegpt4o}, websight~\cite{laurençon2024unlocking}, hateful\_memes~\cite{hatefulmeme}, vision\_flan~\cite{xu2024visionflan}, llava-150K~\cite{liu2023improvedllava}, SoM-llava~\cite{yan2024list}, vsr~\cite{Liu2022VisualSR}
    \item \textbf{OCR, Document and chart understanding (270K)}: DocVQA~\cite{DocVQA}, ChartQA~\cite{ChartQA}, AI2D~\cite{AI2D}, DVQA~\cite{dvqa}, stvqa~\cite{stvqa}, infographicVQA~\cite{mathew2021infographicvqa}, rendered\_text~\cite{renderedtext}, TextVQA~\cite{textvqa}, RobuT Sqa~\cite{zhao2023robut}, HiTab, RobuT wikisql~\cite{zhao2023robut}, vistext, chart2text, arxivQA, hme100k~\cite{yuan2022syntaxaware}
    \item \textbf{Math, science (280K)}: iconqa~\cite{lu2021iconqa}, intergps~\cite{intergps}, tqa~\cite{tqa}, geomverse~\cite{kazemi2023geomverse}, raven~\cite{zhang2019raven}, mathvision~\cite{wang2024measuring}, scienceQA~\cite{ScienceQA}, cambrian-data-engine~\cite{tong2024cambrian1fullyopenvisioncentric}
    \item \textbf{Text-only SFT data (45K)}: gsmk8k~\cite{gsm8k}, slimorca~\cite{OpenOrca}, orca-math-word-problems~\cite{Orca-Math}, Python-Code-23k-ShareGPT~\cite{pythoncode}, lima~\cite{zhou2023lima}
\end{itemize}

\paragraph{Fine-tuning details.} We fine-tune the xGen-MM(BLIP-3)~\cite{Xue2024xGenMMA} using their official training code~\footnote{https://github.com/salesforce/LAVIS/tree/xgen-mm} with 8 H100 GPUs. We adopt the default training configuration as the original BLIP-3 model~\footnote{https://github.com/salesforce/LAVIS/blob/xgen-mm/scripts/example\_finetune\_xgenmmv1-phi3\_mini\_4k\_instruct.sh} and fine-tune our models for one epoch across all experiments. Please refer to the official BLIP-3 codebase for training details.

\section{Instruction data example}
\label{app:gen}
In the current version of \method, we implement $24$ single-image instruction data generators and $14$ multi-image instruction data generators. We provide examples for both single-image instruction data (Table~\ref{tab:single-image-generator}) and multi-image instruction data (Table~\ref{tab:multi-image-generator}).
% \subsection{Single-image instruction data example (Table~\ref{tab:single-image-generator})}
\input{tables/single_image_generator_example}

% \subsection{Multi-image instruction data example (Table~\ref{tab:multi-image-generator})}
\input{tables/multi_image_generator_example}

\section{Raw results of Figure~\ref{fig:dcs} and  Figure~\ref{fig:dcm}}
We provide the raw results of Figure~\ref{fig:dcs} and Figure~\ref{fig:dcm}, i.e., the evaluation results of models that were fine-tuned with DataComp images with our automatic annotation and scene graph generation pipeline (Table~\ref{tab:app-dcs} and Table~\ref{tab:app-dcm}).

\input{tables/dc_single_image_result}
\input{tables/dc_multi_image_result}

%% file: tables/single_image_generator_example.tex
\begin{table*}[h]
  \centering
  \small
  \resizebox{\linewidth}{!}{
    \begin{tabular}{lll} 
    \toprule
    {\bf Task generator} & {\bf Example question} & {\bf Example answer}\\ 
    \midrule

    \multirow{1}{*}{ExistsObjectGenerator}  
    & Tell me what is the number of stop sign that you see? & 1 \\ 
    
    \midrule

    \multirow{1}{*}{MostObjectGenerator}  
    & Determine from boat and sail, which object is the most commonly found? & boat \\ 
    
    \midrule

    \multirow{1}{*}{LeastObjectGenerator}  
    & Among door and drawer, what is the least frequent object? & door \\ 
    
    \midrule

    \multirow{1}{*}{LeftMostObjectGenerator}  
    & Can you tell among jeans, pants, and horses, which object is located on the far left? & pants \\ 
    
    \midrule

    \multirow{1}{*}{RightMostObjectGenerator}  
    & Can you tell among girl, hat, and sign, which object is positioned on the far right? & hat \\ 
    
    \midrule

    \multirow{1}{*}{TopMostObjectGenerator}  
    & Identify which object is located on the most upward, among cloud, floor, and water? & cloud \\ 
    
    \midrule

    \multirow{1}{*}{BottomMostObjectGenerator}  
    & Provide the extreme bottom object among dome, flags, and crosswalk. & crosswalk \\ 
    
    \midrule

    \multirow{1}{*}{ExistsAttributeGenerator}  
    & Tell me the quantity of long kites. & two \\ 
    
    \midrule

    \multirow{1}{*}{AttributeBBoxGenerator}  
    & Can you tell what are the attributes for the kite positioned at region (0.13, 0.26, 0.24, 0.47)? & blue \\ 
    
    \midrule

    \multirow{1}{*}{TypedAttributeBBoxGenerator}  
    & Provide the shape of the hat found at region (0.89, 0.47, 0.96, 0.51). & round \\ 
    
    \midrule

    \multirow{1}{*}{ExistsRelationGenerator}  
    & Can you tell what is the specific relationship between car and bus? & behind and to the left of \\ 
    
    \midrule

    \multirow{1}{*}{RelationBBoxGenerator}  
    & What kind of relationship exists between objects at (0.0, 0.76, 0.83, 1.0) and (0.79, 0.77, 0.99, 1.0)? & to the left of \\ 
    
    \midrule

    \multirow{1}{*}{HeadRelationGenerator}  
    & Which of glass, kitchen, straw, and wine is to the left of post? & straw \\ 
    
    \midrule

    \multirow{1}{*}{SameObjectSegGenerator}  
    & Can you tell the point that is in the same object as (0.83, 0.52) among (0.85, 0.54) and (0.81, 0.54)? & (0.85, 0.54) \\ 
    
    \midrule

    \multirow{1}{*}{DiffObjectSegGenerator}  
    & Identify in the list of (0.4, 0.57) and (0.45, 0.54), which point is in the part of different objects as (0.4, 0.46)? & (0.45, 0.54) \\ 
    
    \midrule

    \multirow{1}{*}{CloserPointGenerator}  
    & Determine in (0.94, 0.9) and (0.9, 0.23), what is closer point to the camera? & (0.9, 0.23) \\ 
    
    \midrule

    \multirow{1}{*}{FartherPointGenerator}  
    & Identify from (0.38, 0.26) and (0.38, 0.83), which point is positioned farther away in depth? & (0.38, 0.26) \\ 
    
    \midrule

    \multirow{1}{*}{CloserObjectGenerator}  
    & Identify what is the nearer object in depth, out of plants and olives? & olives \\ 
    
    \midrule

    \multirow{1}{*}{FartherObjectGenerator}  
    & Tell me out of wheel and man, which object is located farther to the camera? & man \\ 
    
    \midrule

    \multirow{1}{*}{CloserToAnchorObjectGenerator}  
    & In the list of ceiling and shorts, which object is located nearer to the logo in depth? & shorts \\ 
    
    \midrule

    \multirow{1}{*}{FartherToAnchorObjectGenerator}  
    & Among nose and steps, which object is positioned farther away to the dirt in depth? & nose \\ 
    
    \midrule

    \multirow{1}{*}{SceneGraphObjectQAGenerator}  
    & What is the leafy and small object that the word is to the right of? & tree \\ 
    
    \midrule

    \multirow{1}{*}{SceneGraphRelationQAGenerator}  
    & What is the relation from the object, which is behind the empty and wood shelf, to the object, which the large and lying dog is sitting on? & to the left of \\ 
    
    \midrule

    \multirow{1}{*}{SceneGraphAttributeQAGenerator}  
    & What is the color of the stone object that the green arrow is to the left of? & brown \\ 
    
\bottomrule 
\end{tabular}
}  
\caption{Example QA for Single-Image Generator.}
\label{tab:single-image-generator}
\end{table*}

%% file: tables/multi_image_generator_example.tex
\begin{table*}[h]
  \centering
  \small
  \resizebox{\linewidth}{!}{
    \begin{tabular}{lll} 
    \toprule
    {\bf Task generator} & {\bf Example question} & {\bf Example answer}\\ 
    \midrule

    \multirow{1}{*}{HasRelationMultiGenerator}  
    & Determine which image shows that door is to the left of man? & Image 1 \\ 
    
    \midrule

    \multirow{1}{*}{HasNotRelationMultiGenerator}  
    & Tell me in which image tree isn't to the left of street light? & Image 1 \\ 
    
    \midrule

    \multirow{1}{*}{HasObjectMultiGenerator}  
    & Which image shows buildings? & Image 0 \\ 
    
    \midrule

    \multirow{1}{*}{HasNotObjectMultiGenerator}  
    & Tell me which image doesn't have mirror? & Image 0 \\ 
    
    \midrule

    \multirow{1}{*}{HasAttributedObjectMultiGenerator}  
    & Tell me which image contains black tag? & Image 0 \\ 
    
    \midrule

    \multirow{1}{*}{HasNotAttributedObjectMultiGenerator}  
    & Which image doesn't show striped mane? & Image 1 \\ 
    
    \midrule

    \multirow{1}{*}{HasMostObjectMultiGenerator}  
    & Tell me which image shows the most window? & Image 1 \\ 
    
    \midrule

    \multirow{1}{*}{HasLeastObjectMultiGenerator}  
    & Which image shows the least pole? & Image 0 \\ 
    
    \midrule

    \multirow{1}{*}{CommonObjectMultiGenerator}  
    & Identify the objects that are seen in all of these images. & pot \\ 
    
    \midrule

    \multirow{1}{*}{CommonAttributeMultiGenerator}  
    & Identify what is common attribute of kite across these images? & flying \\ 
    
    \midrule

    \multirow{1}{*}{CountObjectMultiGenerator}  
    & Can you tell what is the number of coat in these images? & 2 \\ 
    
    \midrule

    \multirow{1}{*}{CountAttributeObjectMultiGenerator}  
    & Can you tell what is the number of black jacket in these images? & 2 \\ 
    
    \midrule

    \multirow{1}{*}{CompareRelationMultiGenerator}  
    & Determine the difference of the relation between window and windows across these images. & window is to the right of windows in Image 0, to the left of windows in Image 1. \\ 
    
    \midrule

    \multirow{1}{*}{CompareAttributeMultiGenerator}  
    & Determine what differences can be observed in the attributes of kite across these images? & kite is blue in Image 0, yellow and flying in Image 1. \\ 

    \bottomrule
    \end{tabular}
  }
  \caption{Example QA for Multi-Image Generator.}
  \label{tab:multi-image-generator}
\end{table*}

%% file: tables/dc_single_image_result.tex
\begin{table*}[!t]
\centering
\small
\scalebox{0.80}{
\begin{tabular}{ccccccccccc}
\toprule
& \bf Data Ratio & \bf CVB-2D & \bf CVB-3D & \bf SEED & \bf MMB & \bf MME & \bf QBench2 & \bf MMMU & \bf RealWorldQA & \bf Avg. \\
\midrule
\multicolumn{2}{c}{LLaVA-1.5 instruction data~\cite{liu2024visual}} & 58.0 & 61.0 & 66.8 & 66.7 & 63.2 & 46.4 & 36.2 & 54.3 & 56.6 \\
\midrule
\midrule
\multirow{4}{*}{\bf Replacement}
& 5\% & 58.0 & 62.0 & 66.8 & 67.8 & 63.5 & 46.9 & 36.3 & 54.4 & 57.0 \\
\cmidrule{2-11}
& 10\% & 59.0 & 66.0 & 66.9 & 67.6 & 61.3 & 48.1 & 35.4 & 55.6 & 57.5 \\
\cmidrule{2-11}
& 20\% & 58.0 & 66.0 & \textbf{67.0} & \textbf{68.2} & 61.2 & 47.1 & 36.0 & \textbf{56.6} & 57.5 \\
\cmidrule{2-11}
& 50\% & \textbf{60.0} & \textbf{70.0} & 66.2 & 66.6 & \textbf{64.0} & \textbf{49.9} & \textbf{37.4} & 55.4 & \textbf{58.7} \\
\midrule
\midrule
\multirow{4}{*}{\bf Augmentation}
& 5\% & 54.0 & 62.0 & 66.7 & 66.8 & 63.4 & 46.0 & 36.9 & 55.7 & 56.4 \\
\cmidrule{2-11}
& 10\% & 58.0 & 66.0 & \textbf{67.4} & 66.8 & \textbf{64.5} & 45.7 & 35.6 & 56.3 & 57.5 \\
\cmidrule{2-11}
& 20\% & 57.0 & 68.0 & 66.7 & 66.1 & 62.9 & \textbf{49.8} & 37.3 & 55.0 & 57.9 \\
\cmidrule{2-11}
& 50\% & \textbf{60.0} & \textbf{69.0} & 67.2 & \textbf{67.2} & 64.3 & 48.6 & \textbf{37.9} & \textbf{58.2} & \textbf{59.0} \\
\midrule
\end{tabular}
}
\caption{Raw results of Figure~\ref{fig:dcs}, Results of instruction tuning LLaVA-1.5-7B with \dcs, \ie, single-image instruction data generated from DataComp images with our automatic scene graph generation pipeline.}
\label{tab:app-dcs}
\end{table*}

%% file: tables/dc_multi_image_result.tex
% \begin{table*}[!t]
%   \centering
%   \small

%   \scalebox{0.80}{
%     \begin{tabular}{c c c c c c c c c c c} 
%     \toprule
%     \multicolumn{2}{c}{} &  \multicolumn{2}{c}{\bf Multi-image benchmark} &   \multicolumn{6}{c}{\bf Single-image benchmark} \\
%     \cmidrule(lr){3-4}\cmidrule(lr){5-10}
%     & \bf Data Ratio & \bf Mantis-Eval & \bf MMT & \bf SEED & \bf MMB & \bf MME & \bf QBench2 & \bf MMMU & \bf RealWorldQA & \bf Avg. \\
%     \midrule
%     \multicolumn{2}{c}{Mantis instruction data~\cite{jiang2024mantis}} & 54.4 & 52.9 & 68.1 & 72.8 & 58.5 & 70.1 & 44.3 & 51.5 & 59.1 \\
% \midrule
% \midrule
% \multirow{4}{*}{\bf Replacement}
% & 5\% & 57.6 & 53.4 & 68.1 & 71.7 & 57.8 & \textbf{71.5} & 44.4 & 50.9 & 59.4 \\  
% & 10\% & 59.0 & 53.7 & 68.2 & 72.6 & 58.6 & 68.9 & 43.1 & 51.5 & 59.4 \\ 
% & 20\% & 58.5 & 58.6 & \textbf{68.7} & 72.2 & 59.6 & 69.4 & 44.1 & 51.8 & 59.7 \\
% & 50\% & 54.8 & 54.0 & 67.9 & 72.1 & 58.2 & 66.8 & 43.7 & 51.5 & 58.6 \\
% \midrule 
% \midrule
% \multirow{4}{*}{\bf Augmentation}
% & 5\% & 58.1 & 52.5 & 68.0 & 71.8 & 58.4 & 70.1 & 41.3 & \textbf{52.9} & 59.1 \\ 
% & 10\% & \textbf{61.3} & 53.0 & 68.0 & 72.9 & 60.0 & 67.7 & 42.0 & 51.0 & 59.5 \\ 
% & 20\% & 60.4 & 52.8 & 68.5 & 72.4 & \textbf{60.6} & 68.4 & 43.7 & 52.7 & 59.9 \\ 
% & 50\% & 60.4 & 53.7 & 68.1 & \textbf{73.4} & 60.4 & 69.7 & 43.8 & 51.6 & \textbf{60.1} \\
% \midrule

%     \end{tabular}
%   }
%     \caption{Results of instruction tuning Mantis-SigLIP-8B with \vgm, \ie, multi-image instruction data generated from the Visual Genome images and scene graphs.}
%   \label{tab:model-comparison}
% \end{table*}

\begin{table*}[!t]
  \centering
  \small

  \scalebox{0.80}{
    \begin{tabular}{c c c c c c c c c c c} 
    \toprule
    \multicolumn{2}{c}{} &  \multicolumn{2}{c}{\bf Multi-image benchmark} &   \multicolumn{6}{c}{\bf Single-image benchmark} \\
    \cmidrule(lr){3-4}\cmidrule(lr){5-10}
    & \bf Data Ratio & \bf Mantis-Eval & \bf MMT & \bf SEED & \bf MMB & \bf MME & \bf QBench2 & \bf MMMU & \bf RealWorldQA & \bf Avg. \\
    \midrule
    \multicolumn{2}{c}{Mantis instruction data~\cite{jiang2024mantis}} & 54.4 & 52.9 & 68.1 & 72.8 & 58.5 & 70.1 & 44.3 & 51.5 & 59.1 \\
    \midrule
    \midrule
    \multirow{4}{*}{\bf Replacement}
    & 5\% & 52.5 & \textbf{52.3} & \textbf{68.7} & 73.1 & 58.7 & \textbf{70.2} & \textbf{43.7} & 51.2 & 58.8 \\
    & 10\% & \textbf{60.8} & 52.0 & 68.0 & 73.5 & 58.5 & 69.9 & 42.3 & 51.4 & \textbf{59.6} \\
    & 20\% & 55.8 & 51.8 & \textbf{68.7} & \textbf{73.9} & \textbf{59.9} & 68.0 & 42.9 & 52.0 & 59.1 \\
    & 50\% & 54.4 & 51.8 & 67.2 & 71.1 & 59.5 & 65.9 & 41.4 & \textbf{52.6} & 58.0 \\
    \midrule
    \midrule
    \multirow{4}{*}{\bf Augmentation}
    & 5\% & 59.4 & 52.5 & 68.4 & \textbf{73.2} & 58.1 & 70.1 & 43.3 & 49.4 & 59.3 \\
    & 10\% & \textbf{59.9} & \textbf{52.7} & 68.6 & 73.1 & 57.8 & 70.4 & 42.2 & 51.6 & 59.5 \\
    & 20\% & 58.5 & 52.5 & \textbf{68.8} & 72.4 & \textbf{60.5} & \textbf{70.5} & 43.0 & \textbf{52.8} & \textbf{59.9} \\
    & 50\% & 59.0 & \textbf{52.7} & 68.6 & 73.1 & 59.0 & 69.9 & \textbf{43.7} & 51.4 & 59.7 \\
    \midrule
    
    \end{tabular}
  }
    \caption{Raw results of Figure~\ref{fig:dcm}, Results of instruction tuning Mantis-SigLIP-8B with \dcm, \ie, multi-image instruction data generated from DataComp images with our automatic scene graph generation pipeline.}
  \label{tab:app-dcm}
\end{table*}